\PassOptionsToPackage{table}{xcolor}
\documentclass{selfevolagent}

\microtypesetup{expansion=false}
\usepackage{amsmath}
\usepackage{amssymb}
\usepackage{mathtools}
\usepackage{array}
\usepackage{colortbl}
\usepackage{tabularx}
\usepackage{algorithm}
\usepackage{algpseudocode}
\usepackage{longtable}
\usepackage{listings}
\usepackage{fontawesome5}
\DeclareRobustCommand{\corrauth}{\text{\raisebox{-0.12ex}{\scalebox{0.78}{\faEnvelope}}}}
\lstdefinestyle{harnessconfig}{
    basicstyle=\ttfamily\footnotesize,
    columns=fullflexible,
    keepspaces=true,
    breaklines=true,
    breakatwhitespace=false,
    showstringspaces=false,
    aboveskip=0pt,
    belowskip=0pt
}

\providecommand{\Description}[1]{}
\let\cite\citep
\definecolor{resultGroup}{HTML}{F3F5F9}
\definecolor{resultHighlight}{HTML}{E4F0E9}
\definecolor{resultSeparator}{HTML}{7B838A}

\hypersetup{
    pdftitle={Harness-S: Learning from Sibling Harnesses for Generalizable Agent Training},
    pdfauthor={Kangning Zhang, Wenxiang Jiao, Bowen Fang, Shuai Shao, Jiachen Zhu, Rong Shan, Shijian Wang, Hao Wang, Yuan Lu, Weiwen Liu, Weinan Zhang}
}

\title{Harness-S: Learning from Sibling Harnesses for Generalizable Agent Training}
\author[1]{Kangning~Zhang}
\author[2,\corrauth]{Wenxiang~Jiao}
\author[3]{Bowen~Fang}
\author[1]{Shuai~Shao}
\author[1]{Jiachen~Zhu}
\author[1]{Rong~Shan}
\author[4]{Shijian~Wang}
\author[5]{Hao~Wang}
\author[2,\corrauth]{Yuan~Lu}
\author[1,\corrauth]{Weiwen~Liu}
\author[1,\corrauth]{Weinan~Zhang}

\affiliation[1]{Shanghai Jiao Tong University}
\affiliation[2]{Xiaohongshu Inc.}
\affiliation[3]{Zhongguancun Institute of Artificial Intelligence}
\affiliation[4]{Southeast~University}
\affiliation[5]{Carnegie Mellon University}

\contribution[\corrauth]{Corresponding authors}
\metadata[Contact]{\email{zhangkangning@sjtu.edu.cn}, \email{wenxiangjiaonju@gmail.com}, \email{liuww@sjtu.edu.cn}}
\metadata[Code]{\href{https://github.com/DeepExperience/Harness-S}{\nolinkurl{https://github.com/DeepExperience/Harness-S}}}
\metadata[Keywords]{Language model agents, agent harnesses, reinforcement learning, harness complementarity, harness routing}
\abstract{Harness evolution from a common base exhibits \emph{Locality}, \emph{Complementarity}, and \emph{Nonstationarity}: variants repair different failures, and their utility changes as the policy learns.
Rather than seeking a universally superior harness, we introduce \textbf{Harness-S} to selectively rescue failures of a base execution system through \textbf{sibling harnesses} and transfer the resulting capabilities into a single policy that \textbf{remains effective under the original harness}.
\emph{Directional Harness Evolution} develops specialized \textbf{sibling harnesses} along distinct repair directions, forming a reusable \emph{Sibling Harness Pool} that remains fixed during policy training.
\emph{Complementary Harness Rescue Training} invokes this pool only when all initial attempts under the base harness fail.
The same policy generates fresh rollouts under selected auxiliary harnesses, and these trajectories replace the failed group for reinforcement learning.
An \emph{Adaptive Harness Router} selects the Top-$k$ variants using \emph{Harness Behavior Profiles} and voting across multiple LLMs.
\emph{Policy-aware profile re-anchoring} refreshes this evidence as the policy learns.
Across AppWorld, SWE-Gym, and DBBench, the trained policy substantially outperforms policy-only training when returned to the original base harness.
Behavioral analyses and held-out cases support the transfer of harness-elicited behaviors into the policy itself.
Reusing the router for Rank-1 inference provides additional gains at the same executor rollout count.
}

\begin{document}
\maketitle

\section{Introduction}
\label{sec:introduction}

An agent's behavior depends jointly on its model policy and its \emph{harness}: the prompts, tool interfaces, context and memory mechanisms, and execution rules that govern interaction with the environment.
Even with model weights held fixed, the harness shapes the information the policy receives and how its decisions become actions.
ReAct demonstrates the value of interleaving reasoning and action~\cite{react}, while SWE-agent shows how the design of the agent--computer interface affects task execution~\cite{sweagent}.
Recent methods automate the optimization of prompts, workflows, and harness code from execution feedback~\cite{adas,aflow,gepa,metaharness}.
A central question is whether specialized execution support can improve the policy itself, so that its benefits persist under the original harness.

\begin{figure*}[t]
    \centering
    \includegraphics[width=\textwidth]{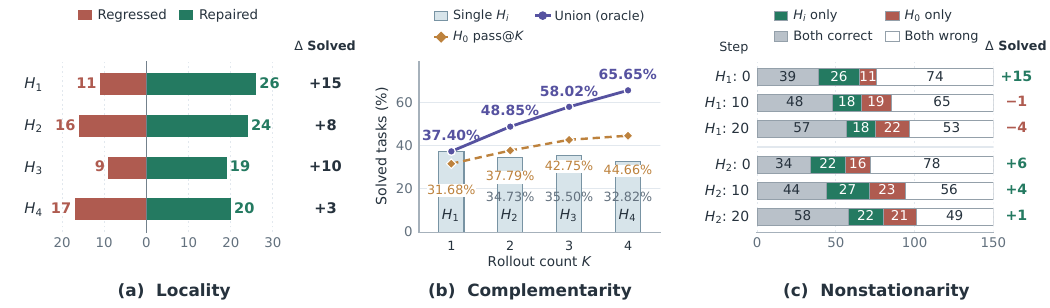}
    \caption{\textbf{Locality, complementarity, and nonstationarity in harness evolution} (Qwen3.5-4B, AppWorld).
    (a) Repaired and regressed tasks relative to $H_0$; $\Delta$ Solved denotes repaired minus regressed tasks.
    (b) Individual success rates (bars), oracle coverage using one rollout under each of $H_1,\ldots,H_K$ (purple), and $H_0$ pass@$K$ using the same number of rollouts (orange).
    (c) Paired outcomes for fixed harnesses as the policy is trained; $\Delta$ Solved is measured against $H_0$ at each checkpoint.}
    \Description{Three horizontally arranged plots for Qwen3.5-4B on AppWorld, with centered titles below each panel. Left, locality: repaired and regressed task counts for H1 through H4 are 26 versus 11, 24 versus 16, 19 versus 9, and 20 versus 17; changes in solved tasks are plus 15, plus 8, plus 10, and plus 3. Middle, complementarity: individual success rates are 37.40, 34.73, 35.50, and 32.82 percent. At rollout counts K equal to 1, 2, 3, and 4, purple cumulative oracle union coverage is 37.40, 48.85, 58.02, and 65.65 percent; the orange dashed H0 pass-at-K curve is 31.68, 37.79, 42.75, and 44.66 percent. Right, nonstationarity: six stacked bars show paired outcomes, colored green for harness-only successes, red for H0-only successes, gray for both correct, and white for both wrong. At steps 0, 10, and 20, H1 changes by plus 15, minus 1, and minus 4 solved tasks, while H2 changes by plus 6, plus 4, and plus 1. The H2 step-10 row contains 27 harness-only successes, 23 H0-only successes, 44 both-correct outcomes, and 56 both-wrong outcomes.}
    \label{fig:harness-specialization}
\end{figure*}

We study harness evolution from a common base $H_0$ within a single execution system, a setting motivated by recent work such as Meta-Harness~\cite{metaharness}, HarnessX~\cite{harnessx}, and Life-Harness~\cite{lifeharness}.
We call the resulting variants \emph{sibling harnesses}: they share the execution foundation of a complete system, such as DeepSeek Harness~\cite{deepseekharness}, while modifying how it supports the policy.
Within this setting, we observe three properties:
\begin{itemize}
    \item \textbf{Locality.}
    Harness edits often repair a subset of failures while introducing regressions on previously solved tasks (Figure~\ref{fig:harness-specialization}a).
    Their effects depend on the failures being targeted, so successive variants develop distinct strengths.
    \item \textbf{Complementarity.}
    Sibling harnesses can recover one another's errors.
    Their oracle union covers more tasks than individual variants or repeated sampling with $H_0$ at the same rollout count (Figure~\ref{fig:harness-specialization}b), showing the benefit of diverse execution behaviors.
    \item \textbf{Nonstationarity.}
    A harness's utility changes as the executing policy learns.
    Its advantage over $H_0$ can become a disadvantage, and the ranking of variants can reverse across policy checkpoints (Figure~\ref{fig:harness-specialization}c).
\end{itemize}

These properties expose two limitations of harness optimization.
\textbf{(i)} Combining local repairs into a universal harness is extremely difficult.
Repairs may impose incompatible execution requirements, and even compatible edits can interfere with one another.
A single accumulated design can therefore lose the strengths of its constituent variants.
\textbf{(ii)} Coupled harness--policy evolution incurs substantial evolution cost and limits harness reuse.
Alternating harness and policy updates~\cite{harnessx,coharness} requires repeated generation and validation of variants as earlier repairs become less useful to the changing policy.
Gains achieved by the evolved pair may also fail to transfer when the policy returns to the base harness $H_0$.
We introduce \textbf{Harness-S} to selectively rescue failures of the base system with specialized sibling harnesses and \emph{transfer the resulting capabilities into a single policy that remains effective under the original $H_0$}.

\textbf{Directional Harness Evolution} first develops variants along distinct repair directions.
Starting from $H_0$, we diagnose failures in execution trajectories and let each child harness target a limited set of related causes.
Joint validation coverage selects the next parent set; all executable, protocol-compliant variants are retained.
The resulting large, reusable \textbf{Sibling Harness Pool} preserves distinct repair specializations and remains fixed throughout policy training.

\textbf{Complementary Harness Rescue Training} uses this pool to expand exploration on tasks the policy fails to solve with $H_0$.
When all initial rollouts for a task fail, the \textbf{Adaptive Harness Router} selects the Top-$k$ auxiliary harnesses, and the same policy generates fresh rollouts under them from the task's initial state.
These rescue rollouts replace the failed group for reinforcement learning, allowing the shared policy to learn from its own actions under complementary execution support.
The objective is to retain the acquired behaviors when the policy returns to $H_0$.

The Adaptive Harness Router matches tasks to sibling harnesses using behavioral evidence and independent votes from multiple LLMs.
\emph{Harness Behavior Profiles} describe the task conditions under which each variant helps or hinders the current policy relative to $H_0$.
To account for nonstationarity, \emph{policy-aware profile re-anchoring} periodically refreshes the profiles from current-policy validation trajectories, keeping routing aligned with the policy as it learns.

Our contributions are fourfold:
\begin{itemize}
    \item We systematically identify \emph{Locality}, \emph{Complementarity}, and \emph{Nonstationarity} in harness evolution and introduce \textbf{Harness-S} to learn complementary sibling capabilities in a single policy that remains effective under the original $H_0$.
    \item We go \emph{beyond the best harness} with \textbf{Directional Harness Evolution}: a pipeline that branches along distinct repair directions and preserves complementary variants in a reusable Sibling Harness Pool.
    \item \textbf{Complementary Harness Rescue Training} restores learning signals when all initial $H_0$ attempts fail. The \textbf{Adaptive Harness Router} selects complementary siblings to recover successful alternatives and within-task reward contrasts for shared-policy optimization.
    \item Across AppWorld, SWE-Gym, and DBBench, Harness-S outperforms policy-only training under $H_0$ at both model sizes. Behavioral analyses, held-out cases, and reduced routing dependence support capability transfer; Rank-1 routing adds gains, and improvements persist under unseen harnesses without further updates.
\end{itemize}

\begin{figure*}[t]
    \centering
    \makebox[\textwidth][c]{%
        \includegraphics[width=1.05\textwidth]{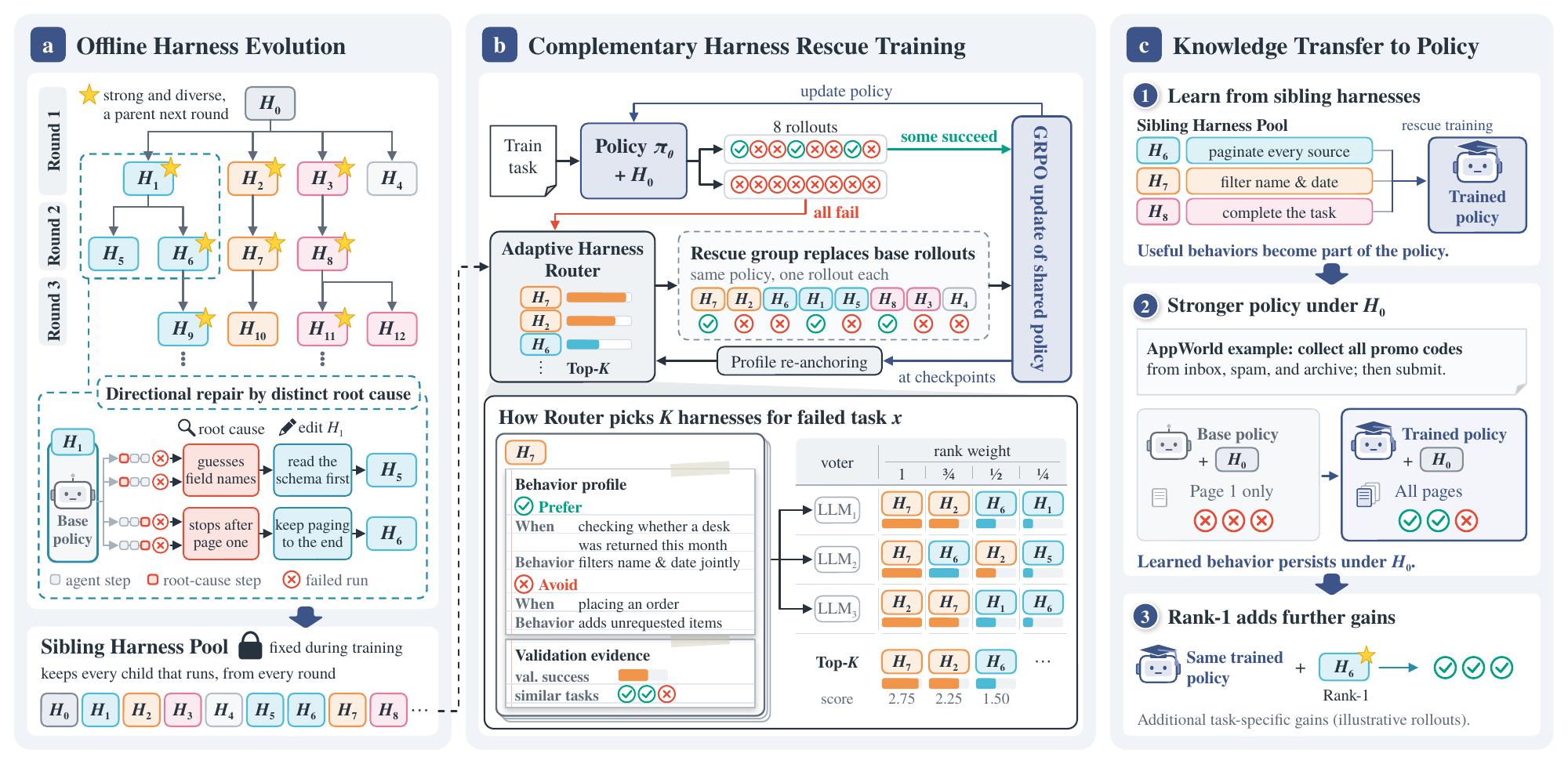}%
    }
    \caption{\textbf{Overview of Harness-S.}
    (a) Directional Harness Evolution develops distinct repairs and retains valid variants in a Sibling Harness Pool, fixed during policy training.
    (b) Complementary Harness Rescue Training routes all-failure tasks to Top-$K$ auxiliary harnesses and updates the shared policy, with periodic profile re-anchoring.
    (c) Rescue training transfers knowledge from sibling harnesses into the shared policy, improving execution under the original $H_0$; Rank-1 routing provides further gains (illustrative AppWorld example).}
    \Description{Three panels depict the Harness-S framework. Left: multiple parent lineages evolve repairs targeting different failure causes, and valid variants accumulate in a fixed sibling pool. Middle: the policy first attempts each task under the base harness. If all attempts fail, the Adaptive Harness Router selects complementary variants using behavior profiles, validation evidence, and votes from multiple LLMs. Fresh rescue rollouts replace the failed group for shared-policy optimization, and profiles are refreshed at policy checkpoints. Right: sibling harnesses contribute pagination, joint filtering, and task-completion behaviors through rescue training. In an illustrative email-retrieval task, the trained policy retains learned behavior under the original base harness, reading all pages across the requested sources. Rank-1 routing provides additional execution support to the same trained policy.}
    \label{fig:method-framework}
\end{figure*}

\section{Related Work}
\label{sec:related-work}

\subsection{Harness Evolution}
Execution design shapes the behavior expressed by a language model, from reasoning and action representations~\cite{react,rewoo,codeact} to interaction interfaces and search~\cite{sweagent,lats}.
Automatic optimization extends this design space through instruction refinement~\cite{opro,textgrad}, evolutionary prompt search~\cite{promptbreeder,evoprompt}, and program compilation or reflective selection~\cite{dspy,gepa}.
Memory, skill libraries, and context adaptation provide another route to improving execution from experience~\cite{reflexion,voyager,awm,ace}.
At the system level, agent design and workflow search optimize executable structures~\cite{adas,aflow}, while code evolution, end-to-end harness optimization, and runtime adaptation develop broader interventions~\cite{dgm,metaharness,lifeharness}.
Harness-S uses directional evolution to retain complementary local repairs in a reusable sibling pool.
The implementations remain fixed during policy training; their behavioral profiles are refreshed to track changes in utility as the shared policy learns.

\subsection{Harness--Policy Optimization}
Reinforcement learning improves agent behavior through outcome feedback~\cite{deepseekmath,ragen,retool,swerl}, supported by executable environments and training interfaces~\cite{swegym,lightning}.
Interactive benchmarks cover web services~\cite{webshop,webarena,taubench}, applications and databases~\cite{appworld,agentbench}, and software engineering~\cite{swebench}.
Joint methods alternate harness and policy updates~\cite{whale,coharness}, evolve matched pairs~\cite{harnessforge}, or learn runtime coordination~\cite{evoharness}.
Model routing~\cite{frugalgpt,routellm,routerdc} and response ensembles~\cite{llmblender,moa} provide related approaches to allocating execution resources.

HarnessBandit schedules one harness per optimizer step using learnability and gradient-based cross-harness transferability signals~\cite{harnessbandit}.
HarnessX couples harness evolution with cross-harness GRPO over a shared replay buffer, grouping trajectories by task across harness and policy versions~\cite{harnessx}.
Harness-S instead asks whether specialized harnesses can selectively rescue failures of a base execution system and transfer the resulting capabilities into a single policy that remains effective under the original harness.
Its allocation unit is a failed task under $H_0$: only all-failure groups trigger fresh rollouts from the current policy across selected siblings in a fixed pool.
Evaluation under $H_0$ tests whether the resulting improvement survives removal of auxiliary execution support.

\subsection{Recursive Self-Improvement}
Recent surveys distinguish self-improvement by whether updates modify model parameters or the surrounding scaffold~\cite{selfimproving_survey}.
Program-level approaches improve optimization code or agent logic that supports subsequent improvements~\cite{stop,godelagent,dgm}.
AIDE$^2$ applies this loop to research agents~\cite{aide2}, while RRSI regularizes harness edits to favor transferable mechanisms~\cite{rrsi}.
SpecForge refines behavioral specifications from execution feedback~\cite{specforge}; self-rewarding models and SEAL instead update parameters using self-generated rewards or adaptation data~\cite{selfrewarding,seal}.
Harness-S uses a fixed sibling pool to rescue base-system failures and transfer the resulting behaviors into one policy; policy updates refresh the evidence used for later rescue allocation.
Its self-improvement loop operates over policy parameters and experience selection, with a fixed optimizer, verifier, and training-time harness pool, rather than recursively redesigning the improvement algorithm.
Appendix~\ref{app:extended-related-work} further discusses agent RL training, knowledge transfer, harness editing, and reusable skills.

\section{Harness-S}
\label{sec:method}

Harness-S uses selective rescue of $H_0$ failures to transfer useful behaviors from sibling harnesses into a shared policy (Figure~\ref{fig:method-framework}).
Directional evolution builds a reusable pool of specialized variants; rescue training uses them to explore tasks that defeat the base harness.
The Adaptive Harness Router selects variants using behavioral evidence that is refreshed as the policy learns, and can be reused at inference.

\subsection{Policy, Harness, and Task Outcome}
\label{sec:formulation}

A policy and its harness jointly determine how an agent executes a task.
For task $x$, the policy $\pi_\theta$ generates a trajectory $\tau\sim p_{\theta,H}(\cdot\mid x)$ under harness $H$, with verifier-defined outcome $R_x(\tau)\in\{0,1\}$.
The harness controls the context and execution rules through which the policy interacts with the environment.
We define its utility for that policy and task as the expected success rate,
\begin{equation}
    u_\theta(H,x)=\mathbb{E}_{\tau\sim p_{\theta,H}(\cdot\mid x)}[R_x(\tau)].
    \label{eq:utility}
\end{equation}

The Sibling Harness Pool $\mathcal H=\{H_0,H_1,\ldots,H_M\}$ contains variants descended from a common base $H_0$.
All variants share the execution foundation, environment interfaces, and task verifier, and use the same policy parameters.
Their relative utility varies across tasks and policy checkpoints, motivating task-specific selection that adapts as the policy learns.

\subsection{Directional Harness Evolution}
\label{sec:offline-evolution}
Starting with $\mathcal P_1=\{H_0\}$, we execute each parent $H\in\mathcal P_g$ on fresh training tasks to derive repair directions $\mathcal D_g(H)$.
Each direction groups related local root causes supported by multiple failed tasks; the editor receives the parent and these causes:
\begin{equation}
    \mathcal C_g=\big\{\operatorname{Edit}(H,d):
    H\in\mathcal P_g,\ d\in\mathcal D_g(H)\big\},
    \label{eq:directional-offspring}
\end{equation}
where $\operatorname{Edit}(H,d)$ modifies the parent's components to address the causes in $d$.

\textbf{Parent selection maximizes joint validation coverage.}
We select at most $B$ parents from $\mathcal C_g^{\mathrm{valid}}$, this round's executable, protocol-compliant children.
Let $\mathcal V(H)$ denote the validation tasks solved by $H$ with the frozen policy:
\begin{equation}
    \mathcal P_{g+1}\in
    \operatorname*{arg\,max}_{\mathcal P\subseteq\mathcal C_g^{\mathrm{valid}},\,1\leq|\mathcal P|\leq B}
    \left|\bigcup\nolimits_{H\in\mathcal P}\mathcal V(H)\right|.
    \label{eq:parent-selection}
\end{equation}
Each task counts once, favoring parents that cover one another's failures.

The Sibling Harness Pool retains all valid children, including those not selected as parents:
\begin{equation}
    \mathcal H_g=\mathcal H_{g-1}\cup\mathcal C_g^{\mathrm{valid}},
    \qquad \mathcal H_0=\{H_0\}.
    \label{eq:pool-archive}
\end{equation}
After $T$ generations, $\mathcal H=\mathcal H_T$ is the fixed pool used for policy training.
Appendices~\ref{app:harness-diversity} and~\ref{app:harness-evolution} report repair diversity and evolution prompts and settings.

\subsection{Complementary Harness Rescue Training}
\label{sec:rescue}

Training uses the pool to revisit tasks that the policy fails to solve under $H_0$.
For each task $x$, the current policy generates $K$ rollouts under $H_0$, forming $\mathcal G_0(x)$.
If all attempts fail, the router selects a set $\mathcal S_K^{(s)}(x)$ of distinct auxiliary harnesses using the latest calibration $s$ (Equation~\eqref{eq:rescue-selection}).
The same policy makes one fresh attempt under each selected harness from the task's initial state, yielding $\mathcal G_R(x)$.
The group retained for training is
\begin{equation}
    \mathcal G(x)=
    \begin{cases}
        \mathcal G_0(x), & \max_{\tau\in\mathcal G_0(x)}R_x(\tau)=1,\\
        \mathcal G_R(x), & \text{otherwise}.
    \end{cases}
    \label{eq:group-replacement}
\end{equation}

\textbf{Complementarity gives the policy alternative paths through difficult tasks.}
Rescue groups containing successes and failures provide reward contrasts absent from the initial attempts.
For a retained group of trajectories $\tau_{x,j}$, GRPO~\cite{deepseekmath} assigns the relative advantage
\begin{equation}
    a_{x,j}=\frac{R_x(\tau_{x,j})-\bar r_x}{\max(\sigma_x,\epsilon)},
    \label{eq:group-advantage}
\end{equation}
Here $\bar r_x$ and $\sigma_x$ are the group's reward mean and standard deviation, and $\epsilon>0$ stabilizes division.
The shared optimizer forms training advantages $A_{x,j}$ from these group-relative rewards and maximizes the clipped token objective
\begin{equation}
    \begin{aligned}
        \rho&=\frac{\pi_\theta(y\mid c_H)}{\pi_{\theta_{\mathrm{old}}}(y\mid c_H)},\\
        \ell(\rho,A)&=\min\!\big\{
        \rho A,\,
        \operatorname{clip}(\rho,1-\eta_-,1+\eta_+)A
        \big\}.
    \end{aligned}
    \label{eq:surrogate}
\end{equation}
Here $y$ is a policy token in its generating harness context $c_H$.
The ratio compares current parameters $\theta$ with pre-update parameters $\theta_{\mathrm{old}}$; $\eta_-,\eta_+$ are the clipping widths.
Only policy tokens enter the objective.
Algorithm~\ref{alg:harness-s} summarizes training; optimizer details appear in Appendix~\ref{app:optimization}.

\begin{algorithm}[t]
    \caption{Complementary Harness Rescue Training}
    \label{alg:harness-s}
    \begin{algorithmic}[1]
        \Require Policy $\pi_\theta$; fixed pool $\mathcal H$; validation tasks
        \State Construct profiles and validation evidence
        \For{each training batch}
            \State Re-anchor profiles if calibration is due
            \For{each task $x$ in the batch}
                \State Sample $K$ rollouts under $H_0$ as $\mathcal G_0(x)$
                \If{all rollouts fail}
                    \State Select $\mathcal S_K^{(s)}(x)$ with the router
                    \State Form $\mathcal G_R(x)$ with one fresh rollout per harness
                \EndIf
                \State Retain $\mathcal G(x)$ using Equation~\eqref{eq:group-replacement}
            \EndFor
            \State Update $\theta$ on retained groups using GRPO
        \EndFor
        \State Re-anchor profiles with the final policy
    \end{algorithmic}
\end{algorithm}

\subsection{Adaptive Harness Router}
\label{sec:router}
\label{sec:profiles}

\textbf{Harness Behavior Profiles} describe when a sibling helps or hinders the current policy:
\begin{equation}
    \begin{aligned}
        P_i^{(s)}&=\big(P_{i,\mathrm{prefer}}^{(s)},P_{i,\mathrm{avoid}}^{(s)}\big),\\
        e&=\{\textsf{When}:w,\ \textsf{Behavior}:b\}.
    \end{aligned}
    \label{eq:profile}
\end{equation}
An LLM groups behavioral differences from paired validation trajectories into entries $e$, where $w$ specifies the task condition and $b$ the behavior relative to $H_0$ (Appendix~\ref{app:harness-profiles}).

\textbf{Voting aggregates independent LLM rankings.}
Each of $V$ voters uses the task, current profiles, validation success rates, and outcomes on similar tasks:
\begin{align}
    q_s(H\mid x)&=\sum_{v=1}^{V}\omega\!\big(\operatorname{rank}_{v,s}(H\mid x)\big),
    \label{eq:voting}\\
    \mathcal S_K^{(s)}(x)&=\operatorname{TopK}_{H\in\mathcal H\setminus\{H_0\}}
    q_s(H\mid x).
    \label{eq:rescue-selection}
\end{align}
Here $\omega$ decreases with rank and is zero for unlisted candidates; the selected $K$ distinct harnesses generate $\mathcal G_R(x)$ in Equation~\eqref{eq:group-replacement}.

\textbf{Policy-aware profile re-anchoring} refreshes the router's evidence at checkpoints $s$.
With policy $\pi_{\theta_s}$ frozen, we collect validation trajectories and outcomes $\mathcal E_i^{(s)}$ under each $H_i$ and recompute
\begin{equation}
    P_i^{(s)}=\operatorname{Profile}\!\big(\mathcal E_i^{(s)},\mathcal E_0^{(s)}\big).
    \label{eq:profile-reanchoring}
\end{equation}
Paired executions share tasks, initial states, and execution limits.
Profiles and validation statistics are refreshed together and used until the next checkpoint; test outcomes are excluded.

\begin{table*}[t]
    \caption{\textbf{Main results on long-horizon agent tasks.} All scores are percentages; higher means are better.
    Bold marks the best mean at each capacity among the compared non-frontier methods.
    $H_{\mathrm{best}}$ is one selected harness held fixed across test tasks; $H_{\mathrm{rank1}}$ is the router's top-ranked harness selected separately for each task.
    DSH denotes DeepSeek Harness.
    Scores are averaged over three independent randomized runs, each with three executor rollouts per task.
    Harness-S entries report mean $\pm$ standard deviation across these runs.}
    \label{tab:main-results}
    \centering
    \begingroup
    \small
    \setlength{\tabcolsep}{2.5pt}
    \renewcommand{\arraystretch}{1.02}
    \begin{tabularx}{\textwidth}{@{}lcc*{6}{>{\centering\arraybackslash}X}@{}}
        \toprule
        \multirow{2}{*}{Model / method} & \multirow{2}{*}{Param.} & \multirow{2}{*}{Test harness} & \multicolumn{2}{c}{AppWorld} & \multicolumn{2}{c}{SWE-Gym} & \multicolumn{2}{c}{DBBench} \\
        \cmidrule(lr){4-5}\cmidrule(lr){6-7}\cmidrule(lr){8-9}
         &  &  & Pass@3 & Avg@3 & Pass@3 & Avg@3 & Pass@3 & Avg@3 \\
        \midrule
        \rowcolor{resultGroup}
        \multicolumn{9}{c}{\textit{Frontier models}} \\
        GPT-6-Sol & --- & $H_0$ & 82.82 & 76.34 & 53.13 & 47.50 & 71.33 & 68.89 \\
        DeepSeek-v4-Flash & --- & $H_0$ & 81.68 & 77.61 & 51.88 & 46.25 & 75.00 & 70.44 \\
        GLM-5.3-Flash & --- & $H_0$ & 80.15 & 65.65 & 46.88 & 43.13 & 76.67 & 72.11 \\
        \midrule[0.35pt]
        \rowcolor{resultGroup}
        \multicolumn{9}{c}{\textit{Base model and policy-only training}} \\
         & 4B &  & 42.62 & 31.93 & 19.38 & 17.50 & 59.00 & 56.89 \\
        \multirow{-2}{*}{Base model} & 9B & \multirow{-2}{*}{$H_0$} & 50.00 & 39.69 & 23.13 & 20.63 & 63.00 & 60.22 \\
        \addlinespace[1pt]
         & 4B &  & 60.31 & 48.98 & 24.38 & 22.50 & 66.67 & 64.00 \\
        \multirow{-2}{*}{GRPO} & 9B & \multirow{-2}{*}{$H_0$} & 64.89 & 53.31 & 28.13 & 25.63 & 69.67 & 67.00 \\
        \addlinespace[1pt]
         & 4B &  & 61.83 & 51.91 & 25.63 & 23.33 & 67.33 & 64.67 \\
        \multirow{-2}{*}{GRPO ($H_0$ rescue)} & 9B & \multirow{-2}{*}{$H_0$} & 65.65 & 54.83 & 29.38 & 26.88 & 70.67 & 67.78 \\
        \addlinespace[1pt]
         & 4B &  & 57.63 & 46.69 & 25.63 & 23.33 & 65.67 & 62.78 \\
        \multirow{-2}{*}{\shortstack[l]{GRPO Mixed Harness\\{\footnotesize (DSH $H_0$ + Claude Code)}}} & 9B & \multirow{-2}{*}{$H_0$} & 62.21 & 50.89 & 29.38 & 26.88 & 68.67 & 65.89 \\
        \addlinespace[1pt]
         & 4B &  & 58.78 & 47.84 & 26.88 & 24.38 & 64.67 & 61.89 \\
        \multirow{-2}{*}{\shortstack[l]{GRPO Mixed Harness\\{\footnotesize (DSH $H_0$ + Top-4)}}} & 9B & \multirow{-2}{*}{$H_0$} & 63.36 & 52.16 & 30.63 & 27.71 & 67.67 & 65.00 \\
        \midrule[0.35pt]
        \rowcolor{resultGroup}
        \multicolumn{9}{c}{\textit{Frozen policy + harness evolution}} \\
         & 4B &  & 48.09 & 36.26 & 21.88 & 20.00 & 62.67 & 60.11 \\
        \multirow{-2}{*}{Meta-Harness} & 9B & \multirow{-2}{*}{$H_{\mathrm{best}}$} & 53.44 & 42.88 & 25.63 & 23.13 & 65.67 & 62.78 \\
        \addlinespace[1pt]
         & 4B &  & 56.87 & 45.42 & 25.00 & 23.75 & 69.33 & 66.78 \\
        \multirow{-2}{*}{Harness-S (HE only)} & 9B & \multirow{-2}{*}{$H_{\mathrm{rank1}}$} & 61.45 & 50.13 & 29.38 & 27.08 & 72.67 & 69.78 \\
        \midrule[0.35pt]
        \rowcolor{resultGroup}
        \multicolumn{9}{c}{\textit{Policy--harness co-evolution}} \\
         & 4B &  & 63.36 & 52.93 & 29.38 & 26.88 & 70.67 & 68.11 \\
        \multirow{-2}{*}{WHALE} & 9B & \multirow{-2}{*}{$H_{\mathrm{best}}$} & 68.70 & 58.02 & 33.75 & 30.63 & 74.00 & 71.11 \\
        \addlinespace[1pt]
         & 4B &  & 59.54 & 49.75 & 27.50 & 25.00 & 69.00 & 66.22 \\
        \multirow{-2}{*}{HarnessForge} & 9B & \multirow{-2}{*}{$H_{\mathrm{best}}$} & 64.89 & 54.83 & 31.25 & 28.54 & 72.00 & 69.22 \\
        \midrule[0.35pt]
        \rowcolor{resultGroup}
        \multicolumn{9}{c}{\textit{Harness-S: offline evolution + policy training}} \\
         & 4B &  & \mbox{67.56{\footnotesize\,$\pm$\,1.87}} & \mbox{56.87{\footnotesize\,$\pm$\,1.47}} & \mbox{38.75{\footnotesize\,$\pm$\,2.22}} & \mbox{28.75{\footnotesize\,$\pm$\,1.73}} & \mbox{73.33{\footnotesize\,$\pm$\,2.36}} & \mbox{71.00{\footnotesize\,$\pm$\,1.84}} \\
        \multirow{-2}{*}{\textbf{Harness-S}} & 9B & \multirow{-2}{*}{$H_0$} & \mbox{72.52{\footnotesize\,$\pm$\,1.65}} & \mbox{62.34{\footnotesize\,$\pm$\,1.28}} & \mbox{40.00{\footnotesize\,$\pm$\,2.04}} & \mbox{32.50{\footnotesize\,$\pm$\,1.51}} & \mbox{76.33{\footnotesize\,$\pm$\,2.13}} & \mbox{73.44{\footnotesize\,$\pm$\,1.67}} \\
        \addlinespace[1pt]
        \rowcolor{resultHighlight}
         & 4B &  & \mbox{\textbf{68.32}{\footnotesize\,$\pm$\,1.74}} & \mbox{\textbf{61.45}{\footnotesize\,$\pm$\,1.35}} & \mbox{\textbf{42.50}{\footnotesize\,$\pm$\,2.04}} & \mbox{\textbf{32.29}{\footnotesize\,$\pm$\,1.64}} & \mbox{\textbf{75.00}{\footnotesize\,$\pm$\,2.23}} & \mbox{\textbf{72.67}{\footnotesize\,$\pm$\,1.75}} \\
        \rowcolor{resultHighlight}
        \multirow{-2}{*}{\textbf{Harness-S}} & 9B & \multirow{-2}{*}{$H_{\mathrm{rank1}}$} & \mbox{\textbf{75.19}{\footnotesize\,$\pm$\,1.56}} & \mbox{\textbf{67.05}{\footnotesize\,$\pm$\,1.20}} & \mbox{\textbf{43.75}{\footnotesize\,$\pm$\,1.84}} & \mbox{\textbf{36.25}{\footnotesize\,$\pm$\,1.45}} & \mbox{\textbf{78.00}{\footnotesize\,$\pm$\,1.96}} & \mbox{\textbf{75.67}{\footnotesize\,$\pm$\,1.55}} \\
        \bottomrule
    \end{tabularx}
    \endgroup
\end{table*}

\subsection{Knowledge Transfer to Policy}
\label{sec:knowledge-transfer}
\label{sec:inference}

The trained policy $\pi_{\theta_*}$ consolidates behaviors from sibling harnesses through rescue updates; these behaviors can persist under the original $H_0$.
The behavioral cases in RQ4 and reduced dependence on auxiliary harnesses in RQ5 (Sections~\ref{sec:rq-behavior} and~\ref{sec:rq-router}) support this transfer.

\textbf{Rank-1 routing adds further gains} to the same trained policy.
After final profile re-anchoring, we select from the full pool, including $H_0$:
\begin{equation}
    \widehat H(x)=\operatorname*{arg\,max}_{H\in\mathcal H}q_*(H\mid x).
    \label{eq:inference-selection}
\end{equation}
Selection precedes execution at a matched rollout count; Appendix~\ref{app:inference-routing} details voting.

\section{Experiments}
\label{sec:experiments}

\begin{figure*}[!t]
    \centering
    \includegraphics[width=\textwidth]{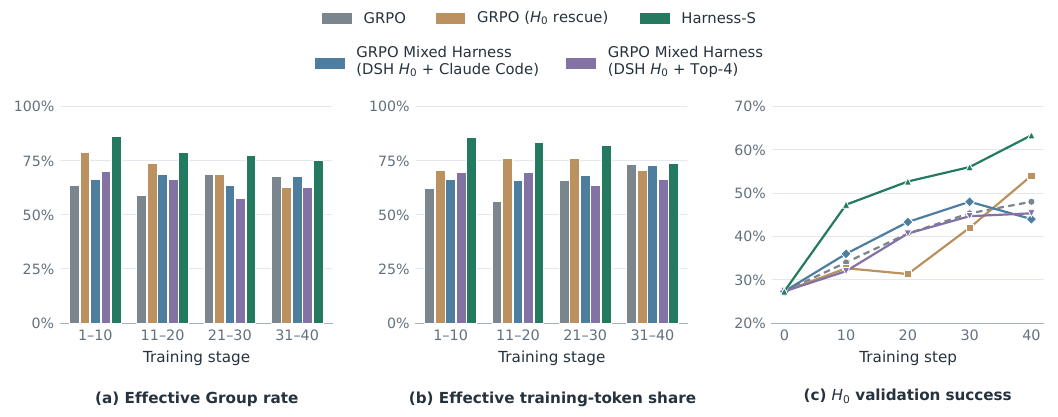}
    \caption{Training signals and policy improvement for Qwen3.5-4B on AppWorld.
    (a) Fraction of effective groups containing both successes and failures.
    (b) Fraction of retained policy tokens belonging to these effective groups.
    Both quantities are aggregated within each ten-step stage.
    (c) Checkpoint success on the same validation set under $H_0$.}
    \Description{Two grouped bar charts and a checkpoint line chart compare GRPO, GRPO with $H_0$ rescue, two fixed mixed-harness GRPO variants, and Harness-S. Bar-chart axes start at zero. Harness-S has the highest Effective Group rate and effective training-token share in every stage and the highest H0 validation success at every checkpoint after step zero. Final validation success is 48.00, 54.00, 44.00, 45.33, and 63.33 percent, respectively.}
    \label{fig:learning-mechanism}
\end{figure*}

We examine whether complementary rescue improves agent performance and transfers useful behaviors into the policy:
\begin{itemize}
    \item RQ1: Does Harness-S improve task success, including under the original $H_0$?
    \item RQ2: Does complementary rescue provide better learning signals than resampling or fixed harness mixtures?
    \item RQ3: How do rescue allocation and profile re-anchoring affect recovery and final performance?
    \item RQ4: Which behaviors change under $H_0$, and do rescue behaviors transfer to untrained instances?
    \item RQ5: Do router ranks predict performance, and how does routing benefit change after training?
    \item RQ6: Do policy improvements transfer to unseen harnesses without further updates?
\end{itemize}

\subsection{Experimental Setup}
\label{sec:experimental-setup}

\textbf{We evaluate Qwen3.5-4B and Qwen3.5-9B on AppWorld~\cite{appworld}, DBBench~\cite{agentbench}, and SWE-Gym~\cite{swegym}, covering application APIs, databases, and software engineering.}
Harness editing and policy optimization use training tasks; validation tasks guide complementary parent selection and profile anchoring.
Pass@3 measures the fraction of tasks solved at least once, and Avg@3 averages success over three rollouts per task.
Main-table scores are means over three independent randomized runs, with across-run standard deviations for Harness-S.
Routed evaluation selects one harness before each task's three attempts.
All experiments use eight NVIDIA H800 GPUs.

\textbf{Our comparisons distinguish policy learning, harness evolution, and joint harness--policy adaptation (Table~\ref{tab:main-results}).}
The GRPO controls include $H_0$ rescue and fixed mixtures with Claude Code or sibling harnesses, all evaluated under $H_0$.
WHALE uses rejection-sampling fine-tuning and HarnessForge adapts branch-specific policies; the GRPO controls share our optimizer.
Meta-Harness, WHALE, and HarnessForge use one selected $H_{\mathrm{best}}$ across test tasks; $H_{\mathrm{rank1}}$ denotes task-specific selection.
Harness-S (HE only) freezes the starting policy, while full Harness-S evaluates the same trained checkpoint under $H_0$ and $H_{\mathrm{rank1}}$ to distinguish retained policy improvement from inference-time support.

\textbf{GRPO with $H_0$ rescue and Harness-S share the all-failure trigger and per-task rescue allocation; their total sampling differs with rescue frequency.}
Dataset splits and implementation protocols appear in Appendices~\ref{app:dataset-splits} and~\ref{app:implementation}, with costs reported in Appendix~\ref{app:computational-efficiency}.

\subsection{Overall Performance (RQ1)}
\label{sec:main-results}

\textbf{Harness-S outperforms policy-only training, frozen-policy harness optimization, and joint adaptation under both $H_0$ and $H_{\mathrm{rank1}}$ (Table~\ref{tab:main-results}).}
The routed 4B policy exceeds 9B GRPO with $H_0$ rescue on every reported metric; the routed 9B policy also exceeds the frontier-model references on DBBench.

\textbf{The gains combine retained policy improvement with better execution support.}
Harness-S outperforms plain GRPO and GRPO with $H_0$ rescue across all benchmark--capacity settings under $H_0$, indicating a benefit beyond additional sampling.
With the policy frozen, Harness-S (HE only) outperforms Meta-Harness, supporting the value of the sibling pool and task-specific selection.
Routing the trained policy further improves both metrics in every setting without increasing the executor rollout count.
On AppWorld 4B, Harness-S outperforms $H_0$ rescue with fewer rollouts and less training GPU time (Appendix~\ref{app:computational-efficiency}).
Its full-pipeline GPU hours and external-model token use fall within the range of the joint-adaptation baselines.

\subsection{Learning Mechanism (RQ2)}
\label{sec:rq-policy}

\textbf{Selective rescue supplies more within-task reward contrast than additional sampling or fixed harness mixtures.}
Figure~\ref{fig:learning-mechanism} compares Qwen3.5-4B on AppWorld under GRPO, $H_0$ rescue, the two fixed mixtures, and Harness-S.
The \emph{Effective Group} (EG) rate measures the fraction of retained groups containing both successes and failures.
Harness-S leads in every stage, reaching 79.38\% over the full run versus 70.94\% for GRPO with $H_0$ rescue and 64.69\% for standard GRPO.
All-failure groups provide no within-task reward contrast; recovering some of these tasks supplies alternatives that GRPO can reinforce.
Both fixed mixtures remain close to standard GRPO, suggesting that harness diversity is more useful when directed toward tasks on which the base group fails.

\textbf{Effective groups also account for a larger share of the retained training tokens.}
The \emph{effective training-token share} counts policy tokens in effective groups relative to all retained policy tokens, accounting for differences in trajectory length.
Harness-S has the highest share throughout training, reaching 80.88\% overall compared with 72.68\% for GRPO with $H_0$ rescue.
Its advantage therefore extends to the tokens used for optimization, accounting for the lengths of successful and unsuccessful trajectories.

\textbf{These training advantages accompany stronger policy performance under $H_0$.}
From the same starting policy, Harness-S leads at every subsequent validation checkpoint and finishes at 63.33\%, compared with 54.00\% for GRPO with $H_0$ rescue and 48.00\% for GRPO.
Both fixed mixtures finish below GRPO despite intermediate gains from the Claude Code mixture.
The results support selective rescue as a source of learning signals whose benefits persist after auxiliary harness support is removed.

\begin{table}[t]
    \centering
    \caption{Rescue success and final test performance on AppWorld (Qwen3.5-4B), in percent. Rule Router uses no Harness Behavior Profiles. Top-8 Fixed uses unchanged profiles; Top-8 Adaptive (Harness-S) uses profile re-anchoring. Bold marks the best result; --- denotes no rescue.}
    \label{tab:rescue-strategies}
    \footnotesize
    \setlength{\tabcolsep}{1.5pt}
    \begin{tabular*}{\columnwidth}{@{\extracolsep{\fill}}lrrrrrr@{}}
        \toprule
        & GRPO & \shortstack{Random\\Harness} & \shortstack{Rule Router\\(w/o Profile)} & \shortstack{Top-4\\$\times2$} & \shortstack{Top-8\\Fixed} & \shortstack{Top-8\\Adaptive} \\
        \midrule
        \rowcolor{resultGroup}
        \multicolumn{7}{@{}c@{}}{\textit{Rescue success by training stage}} \\
        Steps 1--10  & --- & 25.00 & 40.00 & 43.48 & 35.71 & \textbf{52.38} \\
        Steps 11--20 & --- & 18.18 & 26.67 & 43.75 & 24.24 & \textbf{66.67} \\
        Steps 21--30 & --- & 20.00 & 27.78 & 38.89 & 26.67 & \textbf{45.71} \\
        Steps 31--40 & --- & 25.00 & 44.44 & 33.33 & 38.89 & \textbf{50.00} \\
        \midrule
        \rowcolor{resultHighlight}
        \multicolumn{7}{@{}c@{}}{\textit{Final test performance}} \\
        Pass@3 & 60.31 & 62.98 & 66.41 & 65.65 & 64.50 & \textbf{67.56} \\
        Avg@3  & 48.98 & 50.76 & 52.93 & 55.60 & 52.16 & \textbf{56.87} \\
        \bottomrule
    \end{tabular*}
\end{table}

\subsection{Rescue Strategies (RQ3)}
\label{sec:rq-rescue}

\textbf{Task-specific allocation and profile refresh improve rescue at a fixed rollout budget (AppWorld, Qwen3.5-4B; Table~\ref{tab:rescue-strategies}).}
Random Harness samples pool members, while Rule Router (without Profiles) ranks them by success on similar validation tasks.
LLM Router ($4\times2$) allocates two rollouts to each of four selected harnesses; Top-8 allocates one to each of eight, keeping the rescue rollout budget fixed.
The fixed-profile Top-8 variant keeps its profiles unchanged, while Harness-S re-anchors them during training.
Rescue success measures the fraction of rescue-triggering tasks with at least one successful rollout (Appendix~\ref{app:rescue-variants}).

\textbf{Task-specific selection improves rescue success across training stages.}
Both rule-based and LLM-guided strategies outperform random selection throughout training.
Top-8 Adaptive also exceeds LLM Router ($4\times2$), suggesting that broader harness exploration recovers more failures than repeated sampling from fewer harnesses.
Under the same final AppWorld policy, individual siblings repair 11--15 base-harness failures, while the nine analyzed siblings repair 29 in union (Appendix~\ref{sec:rq-pool}).
This is coverage obtained by executing each sibling once; routing must select candidates before observing their outcomes.

\textbf{Adaptive profiles improve routing over a fixed-profile Top-8 variant.}
With the rollout allocation held at one attempt per selected harness, re-anchoring yields higher rescue success in all four stages and improves final Avg@3 by 4.71 percentage points.
This comparison supports updating the router's behavioral evidence as the policy learns.

\textbf{All evaluated rescue strategies improve final test performance over GRPO, with Top-8 Adaptive achieving the highest Pass@3 and Avg@3.}
The remaining strategies show a tradeoff: Rule Router achieves higher Pass@3 than LLM Router ($4\times2$), but lower Avg@3.
Rescue success therefore does not fully determine the downstream ranking, although Top-8 performs best on both recovery during training and final test success.

\begin{table}[t]
    \centering
    \caption{Behavior frequencies under $H_0$ (\%). An LLM labels complete evaluation trajectories; percentages report the fraction exhibiting each behavior. Arrows indicate preferred directions.}
    \label{tab:policy-behaviors}
    \footnotesize
    \setlength{\tabcolsep}{3pt}
    \begin{tabularx}{\columnwidth}{@{}>{\raggedright\arraybackslash}p{0.50\columnwidth}*{3}{>{\centering\arraybackslash}X}@{}}
        \toprule
        Behavior & Base & GRPO & Harness-S \\
        \midrule
        \rowcolor{resultGroup}
        \multicolumn{4}{@{}c@{}}{\textit{AppWorld}} \\
        Schema lookup before first API use $\uparrow$ & 22.9 & 38.2 & 67.6 \\
        Multiple API calls in one execute block $\uparrow$ & 24.8 & 20.5 & 31.5 \\
        Explicit pagination loops $\uparrow$ & 13.8 & 33.5 & 38.3 \\
        Repeated execute code $\downarrow$ & 15.4 & 30.1 & 9.8 \\
        Read-back after the last write $\uparrow$ & 69.2 & 56.9 & 69.4 \\
        \midrule
        \rowcolor{resultGroup}
        \multicolumn{4}{@{}c@{}}{\textit{SWE-Gym}} \\
        Agent-written assertion checks $\uparrow$ & 10.2 & 11.2 & 18.4 \\
        Switch to assertions after blocked tests $\uparrow$ & 15.4 & 13.3 & 48.5 \\
        Fail-edit-retest of the same target $\uparrow$ & 9.2 & 18.2 & 24.1 \\
        Diff inspection after the last edit $\uparrow$ & 16.3 & 23.5 & 27.5 \\
        \bottomrule
    \end{tabularx}
\end{table}

\subsection{Policy Behavior (RQ4)}
\label{sec:rq-behavior}

\textbf{Harness-S changes how the policy gathers evidence and completes actions under $H_0$.}
Table~\ref{tab:policy-behaviors} compares concrete workflows of the base, GRPO, and Harness-S policies.
On AppWorld, Harness-S consults API schemas before first use more often than GRPO and repeats execute code less often, while also increasing pagination, multi-call execution, and read-back after writes.
These shifts suggest greater use of interface information and state verification, rather than a uniform increase in tool activity.
The frequencies describe behavior, not correctness: more calls or checks are useful only when they serve the task.

\textbf{The SWE-Gym trajectories show more explicit verification and recovery from blocked tests.}
Harness-S more often writes assertion checks, revisits a failed target after editing, and inspects the final diff.
When tests are blocked, switching to assertions occurs in 48.5\% of Harness-S cases versus 13.3\% for GRPO.
These patterns describe alternative ways to validate a patch when the standard test workflow is unavailable.

\begin{table}[t]
    \centering
    \caption{Two AppWorld behavior-transfer cases. All eight initial $H_0$ attempts fail in training; the successful rescue enters an applied policy update. Held-out comparisons use $H_0$.}
    \label{tab:behavior-transfer-cases}
    \small
    \setlength{\tabcolsep}{3pt}
    \renewcommand{\arraystretch}{1.12}
    \begin{tabularx}{\columnwidth}{@{}p{0.22\columnwidth}>{\raggedright\arraybackslash}X@{}}
        \toprule
        Stage & Observed execution \\
        \midrule
        \rowcolor{resultGroup}
        \multicolumn{2}{@{}l@{}}{\textbf{Case 1: Set intersection (held-out test instance)}} \\
        Training rescue & H2 selects liked songs that also occur in the library or playlists, then updates and sends the draft. This successful rollout is used at update 10. \\
        GRPO & Uses a union of liked songs and the other sources, sending an overinclusive list. \textbf{Failure.} \\
        Harness-S & Applies the intersection to the new song instances and sends the correct email. \textbf{Success.} \\
        \midrule
        \rowcolor{resultGroup}
        \multicolumn{2}{@{}l@{}}{\textbf{Case 2: Account recovery (held-out validation instance)}} \\
        Training rescue & H5 completes a Venmo account workflow: reads email codes, resets the password, and verifies login. This successful rollout is used at update 3. \\
        GRPO & On the Amazon instance, requests a reset code and reads the inbox but leaves the password unchanged. \textbf{Failure.} \\
        Harness-S & Reads the new codes, resets the Amazon password, and verifies access with the new credentials. \textbf{Success.} \\
        \bottomrule
    \end{tabularx}
\end{table}

\textbf{Successful rescue behaviors also reappear on untrained instances after auxiliary support is removed.}
Table~\ref{tab:behavior-transfer-cases} traces two AppWorld cases from successful training rescues to held-out execution under $H_0$.
The trained policy preserves a set-intersection constraint and completes a feedback-dependent account-recovery workflow where GRPO fails.
The first case concerns applying the same selection rule to new task instances; the second carries a recovery procedure from Venmo to Amazon, including password reset and verification.
They illustrate transfer of both a task constraint and a multi-step procedure across distinct executions.
Both successful rescues entered a policy update, and the corresponding behavior later appears on a different instance under $H_0$.
These cases connect auxiliary execution experience to behavior retained by the policy after the specialized support is removed.

\begin{table}[t]
    \centering
    \caption{AppWorld router rankings before and after Harness-S training (Qwen3.5-4B; \%). Each harness receives three rollouts per task. $H_0$ rate is the task fraction placing $H_0$ at that rank (---: fixed baseline). Bold marks best task scores.}
    \label{tab:router-effectiveness}
    \small
    \setlength{\tabcolsep}{2pt}
    \renewcommand{\arraystretch}{1.1}
    \begin{tabularx}{\columnwidth}{@{}>{\raggedright\arraybackslash}p{0.18\columnwidth}*{6}{>{\centering\arraybackslash}X}@{}}
        \toprule
        \multirow{2}{*}{Harness} & \multicolumn{3}{c}{Before training} & \multicolumn{3}{c}{After training} \\
        \cmidrule(lr){2-4}\cmidrule(lr){5-7}
        & Pass@3 & Avg@3 & $H_0$ rate & Pass@3 & Avg@3 & $H_0$ rate \\
        \midrule
        $H_0$ & 42.62 & 31.93 & --- & 67.56 & 56.87 & --- \\
        \rowcolor{resultHighlight}
        Rank-1 & \textbf{56.87} & \textbf{45.42} & 1.53 & \textbf{68.32} & \textbf{61.45} & 22.52 \\
        Rank-2 & 53.44 & 42.11 & 6.87 & 66.03 & 55.60 & 14.89 \\
        Rank-3 & 50.38 & 39.19 & 4.96 & 64.89 & 54.20 & 13.36 \\
        \bottomrule
    \end{tabularx}
\end{table}

\subsection{Router Effectiveness (RQ5)}
\label{sec:rq-router}

\textbf{The router's ranking predicts execution quality before and after training.}
Table~\ref{tab:router-effectiveness} compares Qwen3.5-4B under $H_0$ and three task-specific router ranks, each evaluated with three rollouts per task.
Both Pass@3 and Avg@3 decrease from Rank-1 to Rank-3 before and after training.
All three ranked choices outperform $H_0$ before training, showing the value of specialized execution support for the initial policy.
The preserved ordering indicates that routing remains useful as the policy improves.

\textbf{Training makes $H_0$ a more frequent leading choice.}
The fraction of tasks placing $H_0$ at Rank-1 rises from 1.53\% to 22.52\%; its selection rates at Rank-2 and Rank-3 also increase.
The largest increase occurs at Rank-1, showing that $H_0$ more often becomes the router's preferred execution choice.
This increased preference for the base harness is consistent with reduced dependence on auxiliary execution support.

\textbf{The trained policy retains much of the routed advantage under the original harness.}
The Rank-1 margin over $H_0$ shrinks from 14.21 to 0.76 points in Pass@3 and from 13.49 to 4.58 points in Avg@3.
The trained policy under $H_0$ also overtakes Rank-2 and Rank-3 on both metrics.
Together with the cases in Section~\ref{sec:rq-behavior}, these performance and routing changes support the internalization of behaviors elicited during rescue.
Rank-1 retains an advantage in the reported means; its larger remaining Avg@3 margin suggests that task-specific execution support still improves consistency across attempts.

\begin{figure}[t]
    \centering
    \includegraphics[width=0.75\linewidth]{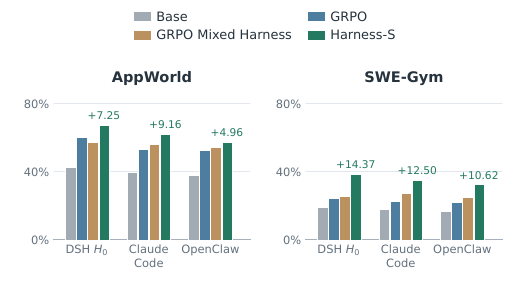}
    \caption{Cross-harness Pass@3 of Qwen3.5-4B. Mixed-harness GRPO uses DSH $H_0$ and Claude Code. Annotations show Harness-S gains over GRPO in percentage points.}
    \Description{Two grouped bar charts compare Base, GRPO, mixed-harness GRPO, and Harness-S on AppWorld and SWE-Gym under DSH H0, Claude Code, and OpenClaw. Only Pass at 3 is shown. Harness-S scores highest in each setting; annotations report its gain over GRPO.}
    \label{fig:harness-generalization}
\end{figure}

\subsection{Cross-Harness Generalization (RQ6)}
\label{sec:rq-generalization}

\textbf{Harness-S retains its performance advantage under external execution harnesses without further policy updates.}
Figure~\ref{fig:harness-generalization} compares the Qwen3.5-4B policies on AppWorld and SWE-Gym under DSH $H_0$, Claude Code, and OpenClaw.
Harness-S achieves the highest Pass@3 across all evaluated benchmarks and execution harnesses.

\textbf{Cross-harness transfer preserves the gains without eliminating sensitivity to the execution interface.}
Harness-S also exceeds mixed-harness GRPO under Claude Code and OpenClaw, indicating that fixed mixing is insufficient to match its transfer performance.
Harness-S nevertheless scores below DSH $H_0$ on both external harnesses and benchmarks.
The results suggest that complementary rescue develops behaviors useful beyond the training harness, while execution support continues to influence how reliably those behaviors succeed.

\FloatBarrier
\section{Conclusion and Future Directions}
\label{sec:conclusion}

Harness-S trains a shared policy to retain capabilities elicited by complementary sibling harnesses under the original $H_0$.
Directional evolution preserves distinct repair strategies, while selective rescue recovers learning signals from base-harness failures.
Across AppWorld, SWE-Gym, and DBBench, gains persist under $H_0$; behavioral analyses support transfer, and Rank-1 routing improves inference.
Future work will study lasting transfer and efficient pool reuse across policies.
We will study how independently developed harnesses exchange task state across execution interfaces.
Harness Scaling experiments will measure how additional repair diversity affects policy gains and calibration cost.

\clearpage
\bibliographystyle{assets/plainnat}
\bibliography{references}

\clearpage
\beginappendix
\section{Dataset Splits}
\label{app:dataset-splits}

Table~\ref{tab:dataset-splits} lists task-instance counts.
Harness editing and policy optimization use training tasks; parent selection and profile anchoring use validation tasks.
Test tasks are reserved for final evaluation.

\begin{table}[htbp]
    \caption{\textbf{Dataset splits.} Number of task instances in each benchmark.}
    \label{tab:dataset-splits}
    \centering
    \begingroup
    \setlength{\tabcolsep}{3pt}
    \begin{tabularx}{\columnwidth}{@{}l*{4}{>{\centering\arraybackslash}X}@{}}
        \toprule
        Benchmark & Train & Validation & Test & Total \\
        \midrule
        AppWorld & 320 & 150 & 262 & 732 \\
        DBBench & 1,600 & 200 & 300 & 2,100 \\
        SWE-Gym & 240 & 100 & 160 & 500 \\
        \bottomrule
    \end{tabularx}
    \endgroup
\end{table}

\section{Shared Policy Optimization}
\label{app:optimization}

The baseline and Harness-S share the same optimizer, so their learning objectives differ only through the trajectories retained for each task.
Each retained group contains $G=K=8$ trajectories.
Equation~\eqref{eq:group-advantage} first centers rewards within the task and normalizes them using population standard deviation and $\epsilon=10^{-6}$.
For a batch $\mathcal B$, let $\mathcal T_{x,j}$ contain the valid policy-token indices and $N_x=\sum_j|\mathcal T_{x,j}|$.
Using the generating context $c_{H_{x,j},\ell}$, the PPO-style objective~\cite{ppo} is
\begin{align}
    \rho_{x,j,\ell}(\theta)
    &=\frac{\pi_\theta(y_{x,j,\ell}\mid c_{H_{x,j},\ell})}
    {\pi_{\theta_{\mathrm{old}}}(y_{x,j,\ell}\mid c_{H_{x,j},\ell})},
    \label{eq:policy-ratio}\\
    \ell_{x,j,\ell}(\theta)
    &=\ell\!\big(\rho_{x,j,\ell}(\theta),A_{x,j}\big),\\
    J(\theta)&=\frac{1}{|\mathcal B|}\sum_{x\in\mathcal B}\frac{1}{N_x}
    \sum_{j=1}^{G}\sum_{\ell\in\mathcal T_{x,j}}
    w_{x,j,\ell}\,\ell_{x,j,\ell}(\theta).
    \label{eq:objective}
\end{align}
Here $\ell$ is the clipped objective in Equation~\eqref{eq:surrogate}.
The loss averages over valid policy tokens within each task, then equally over tasks in the batch; trajectories of different lengths therefore need not receive equal weight.
Harness-injected instructions and tool observations are excluded from the token loss.
Each auxiliary trajectory retains its generating context when computing the probability ratio.
Tasks with $N_x=0$ contribute zero.

The rollout and training backends can assign different probabilities to the same sampled token.
For recorded rollout token probability $q_{x,j,\ell}$, the detached correction is
\begin{equation}
    w_{x,j,\ell}=\operatorname{clip}\!\left(
    \frac{\pi_{\theta_{\mathrm{old}}}(y_{x,j,\ell}\mid c_{H_{x,j},\ell})}
    {q_{x,j,\ell}},\,0,\,2\right).
\end{equation}
The detached weight $w$ multiplies the token objective without receiving gradients.
Policy-ratio clipping uses $\eta_-=0.20$ and $\eta_+=0.28$, with no reference-model KL term.

\section{Implementation of Comparison Variants}
\label{app:implementation}

The adaptations start from the size-matched Qwen3.5 policy and common base harness.
Environment interfaces and task verifiers remain unchanged; each method retains its own search organization and policy-update objective.
$H_{\mathrm{best}}$ is fixed across test tasks, whereas $H_{\mathrm{rank1}}$ is selected separately for each task.
The base model denotes the unmodified starting policy; frontier models evaluated under $H_0$ serve as additional references.

\textbf{Evaluation protocol.}
\label{app:evaluation-protocol}
Within each run, each test task receives three executor rollouts.
Pass@3 measures the fraction of tasks with at least one successful rollout, while Avg@3 averages success over the three attempts.
Main-table scores are means over three independent randomized runs; the reported Harness-S standard deviations are computed across runs.
For the router-rank comparison, Rank-$r$ denotes the candidate at position $r$ for each task, evaluated separately with the same rollout count.

\textbf{GRPO and $H_0$ rescue.}
Both use the shared optimizer in Appendix~\ref{app:optimization} and execute under $H_0$.
Plain GRPO retains each initial group; the rescue control replaces an all-failure group with fresh $H_0$ rollouts using the same per-task allocation as Harness-S.
Total sampling depends on rescue frequency, which differs between policies, and includes both discarded base attempts and rescue rollouts.

\textbf{Rescue allocation variants.}
\label{app:rescue-variants}
All rescue groups contain eight rollouts.
Random Harness samples pool members.
Rule Router ranks candidates by success on similar validation tasks, without using profiles.
LLM Router ($4\times2$) allocates two rollouts to each of four selected harnesses; Top-8 allocates one to each of eight.
The fixed-profile Top-8 variant keeps its profiles unchanged, while Harness-S refreshes them at calibration checkpoints.
Rescue success is the fraction of rescue-triggering tasks for which at least one fresh attempt succeeds.

\textbf{Fixed mixed-harness GRPO.}
The Claude Code mixture allocates four rollouts to DSH $H_0$ and four to Claude Code.
The Top-4 mixture combines DSH $H_0$ with four selected siblings throughout training.
The same mixture is used across training tasks, without waiting for an all-failure base group.
These controls expose one shared policy to diverse harness contexts throughout training.
Only policy parameters are updated, and both policies are evaluated under $H_0$.

\textbf{Meta-Harness.}
With the policy frozen, the proposer uses previous implementations, scores, and trajectories to revise harness code~\cite{metaharness}.
The adaptation searches within the common execution framework, preserving environment interfaces and the task verifier.
Candidates are checked for compatibility and evaluated on training tasks.
A single selected $H_{\mathrm{best}}$ is used across test tasks.

\textbf{WHALE.}
The adaptation alternates Meta-Harness-style search with rejection-sampling fine-tuning on successful trajectories~\cite{whale}.
Successful trajectories supply the policy update, after which harness search proceeds with the updated policy.
This preserves the alternation between model adaptation and execution-support optimization.
Final evaluation uses the selected policy and its fixed $H_{\mathrm{best}}$.

\textbf{HarnessForge.}
Failure analysis guides harness changes, and successful search trajectories train the corresponding branch policy~\cite{harnessforge}.
Branches retain separate policy adaptations and the provenance of their training trajectories.
As recorded in Table~\ref{tab:pipeline-efficiency}, training reuses successful search trajectories, so zero additional training rollouts does not mean that the policy remains frozen.
The final selected pair executes with a fixed $H_{\mathrm{best}}$.

\textbf{Pool and evaluation.}
\label{app:routing-pool}
The pool contains $H_0$ and 19 auxiliary harnesses; rescue selects eight auxiliaries and inference ranks all 20.
The HE-only variant freezes the starting policy; the two trained variants share a checkpoint and differ only in test-harness selection.
The nine siblings analyzed in Appendix~\ref{app:harness-diversity} define the scope of that calibration analysis, not the size of the routing pool.

\textbf{Profile refresh.}
For AppWorld 4B, profiles are built before training, refreshed every ten updates, and re-anchored with the final policy.
Each calibration freezes the current policy for paired validation execution.
Routing uses the latest profiles between checkpoints, with all harness implementations fixed.

\textbf{Inference voting.}
\label{app:inference-routing}
Multiple LLMs independently rank candidates using the task, current profiles, validation success rates, and outcomes on similar validation tasks.
Each ballot ranks four distinct candidates with weights $1$, $3/4$, $1/2$, and $1/4$.
Unlisted candidates receive zero weight.
Weights are summed across ballots; ties use validation success, then harness ID.
Candidate order varies across calls.
The base harness uses its execution description and validation evidence.
Each selected harness is fixed across the three evaluation rollouts for its task.
The ranking is completed before any task outcome is observed, and failed attempts do not trigger a new inference-time selection.

\section{Sibling Harness Edits and Behavior Profiles}
\label{app:harness-profiles}

Table~\ref{tab:harness-profile-examples} pairs three concrete configuration edits with favorable behavior-profile entries from the step-40 AppWorld calibration of Qwen3.5-4B.
The policy is frozen while each sibling is compared with $H_0$ under matched validation tasks, execution limits, and task seeds.
These examples use the paired calibration protocol in Appendix~\ref{app:harness-diversity}.

Harness Behavior Profiles describe execution differences observed in complete paired trajectories.
The \emph{When} field identifies the task conditions, and \emph{Behavior} records the corresponding execution pattern.
Entries are organized into prefer and avoid lists; repeated observations are consolidated by task condition and behavior, while useful observations from individual tasks are retained.
The table presents English translations of selected prefer entries alongside core excerpts of the recorded configuration changes.

The edits add guidance to the shared base execution system.
A \texttt{null} entry means that $H_0$ has no corresponding additional rule or hook; its base instructions remain in place.
The pagination and completion hooks append reminders, while the intersection rule supplies an instruction through \texttt{action\_guidance}.
Each profile describes execution under the full sibling configuration.
The $S_7$ and $S_8$ examples concern the same validation task, highlighting constraint filtering and final submission, respectively.

\begingroup
\small
\setlength{\tabcolsep}{4pt}
\renewcommand{\arraystretch}{1.08}
\begin{longtable}{@{}p{0.18\textwidth}p{\dimexpr0.82\textwidth-2\tabcolsep\relax}@{}}
    \caption{Three sibling configuration edits and favorable behavior-profile excerpts on AppWorld. Each paired example succeeds with the sibling and fails under $H_0$. Code shows the changed entry; \texttt{[...]} marks omitted rule text.}
    \label{tab:harness-profile-examples}\\
    \toprule
    \endfirsthead
    \multicolumn{2}{@{}l@{}}{\textbf{Table \thetable\ (continued)}}\\
    \toprule
    \endhead
    \bottomrule
    \endfoot
    \rowcolor{resultGroup}
    \multicolumn{2}{@{}l@{}}{\textbf{Case 1: $S_6$ --- complete pagination across sources}}\\*
    \multicolumn{2}{@{}p{\textwidth}@{}}{\textbf{Edit:} add an after-tool reminder to continue pagination until explicit exhaustion.}\\*
    \multicolumn{2}{@{}p{\textwidth}@{}}{\texttt{state.hooks.pagination-completion-guidance}}\\*
    \textit{Before ($H_0$)} & \textit{After ($S_6$)}\\*
    \begin{minipage}[t]{\linewidth}
\begin{lstlisting}[style=harnessconfig]
null
\end{lstlisting}\end{minipage}
    & \begin{minipage}[t]{\linewidth}
\begin{lstlisting}[style=harnessconfig]
{
  "action": "warn",
  "condition": "pagination_present",
  "event": "after_tool",
  "message": "Pagination metadata is present. Continue only until an empty or short page or an authoritative total; do not treat the first page as complete.",
  "parameters": {}
}
\end{lstlisting}\end{minipage}\\*
    \textbf{When} & Aggregating information across inbox, spam, and archive.\\*
    \textbf{Behavior} & Paginate through all specified sources, merge the results, and remove duplicates. The paired $H_0$ execution attempts a single-pass match and omits information from some sources.\\
    \addlinespace[6pt]
    \midrule
    \rowcolor{resultGroup}
    \multicolumn{2}{@{}l@{}}{\textbf{Case 2: $S_7$ --- joint filtering of target records}}\\*
    \multicolumn{2}{@{}p{\textwidth}@{}}{\textbf{Edit:} add an action-guidance rule requiring the intersection of task conditions.}\\*
    \multicolumn{2}{@{}p{\textwidth}@{}}{\texttt{state.prompt\_rules.reconcile-before-side-effect}}\\*
    \textit{Before ($H_0$)} & \textit{After ($S_7$)}\\*
    \begin{minipage}[t]{\linewidth}
\begin{lstlisting}[style=harnessconfig]
null
\end{lstlisting}\end{minipage}
    & \begin{minipage}[t]{\linewidth}
\begin{lstlisting}[style=harnessconfig]
{
  "slot": "action_guidance",
  "text": "[...] If the request combines conditions, fetch each condition separately and compute the intersection; do not broaden an item to its containing collection. [...]"
}
\end{lstlisting}\end{minipage}\\*
    \textbf{When} & Finding office desks returned in the current month.\\*
    \textbf{Behavior} & Retrieve all return records, then jointly filter for product names containing \texttt{desk} and a \texttt{returned\_at} date within the current month. The selected records satisfy both the product and time constraints.\\
    \addlinespace[6pt]
    \midrule
    \rowcolor{resultGroup}
    \multicolumn{2}{@{}l@{}}{\textbf{Case 3: $S_8$ --- reserving execution budget for submission}}\\*
    \multicolumn{2}{@{}p{\textwidth}@{}}{\textbf{Edit:} add a low-budget reminder to finish necessary checks and reserve a step for completion.}\\*
    \multicolumn{2}{@{}p{\textwidth}@{}}{\texttt{state.hooks.low-budget-completion}}\\*
    \textit{Before ($H_0$)} & \textit{After ($S_8$)}\\*
    \begin{minipage}[t]{\linewidth}
\begin{lstlisting}[style=harnessconfig]
null
\end{lstlisting}\end{minipage}
    & \begin{minipage}[t]{\linewidth}
\begin{lstlisting}[style=harnessconfig]
{
  "action": "warn",
  "condition": "remaining_steps_lte",
  "event": "after_tool",
  "message": "Execution budget is low. Finish only the required verification and reserve one step for complete_task; avoid unrelated exploration.",
  "parameters": {
    "threshold": 3
  }
}
\end{lstlisting}\end{minipage}\\*
    \textbf{When} & Task-specific filtering and verification are complete, and the final answer must be submitted through the dedicated API.\\*
    \textbf{Behavior} & Finish verification and submit the answer with \texttt{complete\_task}. The paired $H_0$ trajectory reaches a conclusion but ends before submitting it.\\
\end{longtable}
\endgroup

\section{Sibling Harness Diversity}
\label{app:harness-diversity}
\label{sec:rq-pool}

\begin{figure}[t]
    \centering
    \includegraphics[width=0.70\linewidth]{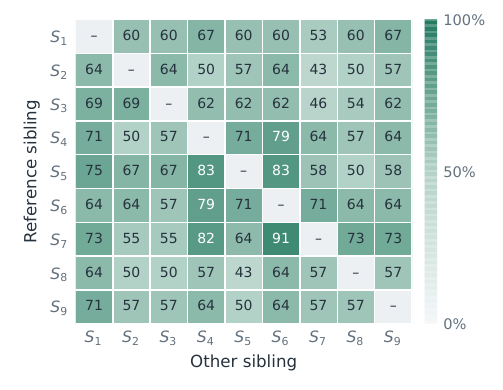}
    \caption{Shared repair coverage. Each cell gives the percentage of the row sibling's repairs also supplied by the column sibling; diagonal entries are omitted.}
    \Description{A nine-by-nine heatmap of directed repair overlap for siblings S1 through S9 under a fixed AppWorld policy. Off-diagonal entries show the fraction of each row's repaired base-harness failures also repaired by the column sibling.}
    \label{fig:sibling-diversity}
\end{figure}

We compare nine siblings using the final AppWorld Qwen3.5-4B policy and one greedy rollout per validation task.
Each harness uses the same task initialization and execution limits.
One task has an invalid paired outcome and is excluded from every sibling, leaving 149 shared tasks.
For each sibling, $R_i$ contains failures of $H_0$ repaired by $S_i$.
Figure~\ref{fig:sibling-diversity} shows the directed overlap $|R_i\cap R_j|/|R_i|$, averaging 62.47\% across distinct ordered pairs.
The denominator is the row sibling's repair set, so the matrix is directional: overlap can differ when the row and column are exchanged.
High overlap indicates that another sibling supplies many of the same repairs; lower overlap identifies task coverage not shared by that pair.
This comparison concerns repairs of $H_0$ failures, rather than overall success across all validation tasks.
The incomplete overlap supports retaining different repair directions among these nine analyzed siblings; the routing pool contains 20 harnesses.
Individually, the analyzed siblings repair 11--15 base-harness failures, while their union repairs 29.
This union is oracle coverage obtained by executing each sibling once; routing must select candidates before observing their task outcomes.

\section{Directional Harness Evolution: Prompts and Settings}
\label{app:harness-evolution}
\label{app:evolution-settings}

The AppWorld search freezes Qwen3.5-4B Base and uses the 320 training tasks for failure analysis, candidate editing, and local tests.
Validation tasks guide complementary parent selection and profile anchoring.
Table~\ref{tab:evolution-settings} summarizes the configuration.
The analysis prompt requires every valid failure to be grouped by root cause or marked unresolved.
The evidence determines how many groups are formed.
Each candidate then targets a limited set of related causes supported by multiple failed tasks.
Component choices follow this diagnosis: edits can alter prompts, skills, context, memory, hooks, or the tool view.
Other candidate briefs provide context for seeking different mechanisms, triggers, or recovery strategies.

\begin{table}[t]
    \centering
    \caption{Selected settings for the AppWorld directional evolution run. Proposal counts are budgets, not counts of accepted variants.}
    \label{tab:evolution-settings}
    \small
    \setlength{\tabcolsep}{4pt}
    \renewcommand{\arraystretch}{1.1}
    \begin{tabularx}{\columnwidth}{@{}lX@{}}
        \toprule
        Setting & Value \\
        \midrule
        Search & 3 rounds; proposal budgets 7 / 8 / 8; 4 parent slots \\
        Development tasks & Sampled from the 320-task training set \\
        Repair direction & 2--3 related root causes per candidate; at least 2 distinct failed tasks per cause \\
        Local evaluation & At most 30 tasks, including at most 10 successful controls \\
        Revision budget & Initial candidate plus at most one revision \\
        Harness editor & GLM-5.3; temperature 0.6; thinking enabled with high reasoning effort \\
        Policy sampling & Greedy decoding; one rollout per task; seed 42 \\
        Execution limits & 32 model calls and 30 environment actions; 65,532 context tokens and 8,192 output tokens per call \\
        Editable components & Prompt, Skill, Context, Memory, Hooks, Tool View \\
        \bottomrule
    \end{tabularx}
\end{table}

Local evaluation tests whether an edit repairs the targeted failures while preserving behavior on successful controls.
Each candidate can submit an initial measured version and one revision within the task budget in Table~\ref{tab:evolution-settings}.
The editing agent submits an artifact containing the candidate state and its self-review, then requests static checks or measured rollouts before final submission.
The execution host supplies these outcomes; candidate self-assessment does not determine success.
Newly generated valid variants compete for $B=4$ parent slots under the validation-coverage objective in Equation~\eqref{eq:parent-selection}.
Selected parents are paired with locally diagnosed root causes for the next round of editing, while the archive retains all executable, protocol-compliant variants.
The resulting implementations remain fixed during policy training; only their profiles and allocation are updated.

\textbf{Prompt excerpts.}
\label{app:evolution-prompts}
Stages receive the parent harness, complete trajectories, and scoring diagnostics; hidden scoring answers are not supplied to the executor policy.
The saved instructions include:
\begin{itemize}
    \raggedright
    \item \textbf{Attribution:} ``For EACH failure, follow the earliest decision or tool result explaining the failed check, then inspect the later effect.''
    \item \textbf{Planning:} ``Choose among all six editable components according to the diagnosed cause; do not default to prompt/skill edits.''
    \item \textbf{Submission:} ``Submit must summarize observed gains, regressions and unresolved causes; never claim unrun tests.''
\end{itemize}

\section{Computational Efficiency}
\label{app:computational-efficiency}
\label{sec:rq-efficiency}

All experiments use eight NVIDIA H800 GPUs; external-model usage is reported separately in API tokens.
The two tables distinguish policy-training expenditure from the complete search-and-training pipeline.

\begin{table}[!htbp]
    \centering
    \caption{GRPO training cost and $H_0$ test Avg@3 on AppWorld (Qwen3.5-4B). All settings use 40 updates. Parentheses give rollout counts relative to plain $H_0$ GRPO.}
    \label{tab:training-efficiency}
    \small
    \setlength{\tabcolsep}{3pt}
    \renewcommand{\arraystretch}{1.12}
    \begin{tabularx}{\columnwidth}{@{}Xrrr@{}}
        \toprule
        Setting & \shortstack{Training\\rollouts} & \shortstack{Training\\GPU$\cdot$h} & \shortstack{$H_0$ test\\Avg@3 (\%)} \\
        \midrule
        Plain $H_0$ GRPO & \shortstack[r]{2,560\\(100.00\%)} & 50.58 & 48.98 \\
        GRPO ($H_0$ rescue) & \shortstack[r]{3,616\\(141.25\%)} & 97.77 & 51.91 \\
        DSH $H_0$ + Top-4 & \shortstack[r]{2,560\\(100.00\%)} & 58.47 & 47.84 \\
        DSH $H_0$ + Claude Code & \shortstack[r]{2,560\\(100.00\%)} & 50.03 & 46.69 \\
        \rowcolor{resultHighlight}
        Harness-S / Top-8 & \shortstack[r]{3,032\\(118.44\%)} & 67.19 & \textbf{56.87} \\
        \bottomrule
    \end{tabularx}
\end{table}

\begin{table}[!htbp]
    \centering
    \caption{Full-pipeline resources for harness search and policy adaptation. Training rollouts count additional samples; HarnessForge reuses successful search trajectories. API tokens are in millions (M).}
    \label{tab:pipeline-efficiency}
    \small
    \setlength{\tabcolsep}{3pt}
    \renewcommand{\arraystretch}{1.12}
    \begin{tabularx}{\columnwidth}{@{}Xrrrr@{}}
        \toprule
        Method & \shortstack{Search\\rollouts} & \shortstack{Training\\rollouts} & GPU$\cdot$h & \shortstack{API\\tokens (M)} \\
        \midrule
        Meta-Harness & 8,000 & 0 & 35.24 & 4.89 \\
       WHALE & 8,640 & 2,560 & 95.84 & 9.80 \\
        HarnessForge & 5,680 & 0 & 65.39 & 24.40 \\
        \rowcolor{resultHighlight}
        Harness-S & \textbf{4,010} & 3,032 & 75.79 & 18.69 \\
        \bottomrule
    \end{tabularx}
\end{table}

\textbf{Training costs include unsuccessful attempts as well as retained trajectories.}
All variants in Table~\ref{tab:training-efficiency} use 40 updates, but the rescue methods incur additional rollouts when the initial $H_0$ group fails.
Harness-S improves $H_0$ Avg@3 by 7.89 points over plain GRPO with 18.44\% more rollouts.
It also outperforms $H_0$ rescue with fewer rescue attempts and lower GPU time.
The fixed mixtures match the plain-GRPO rollout count but do not obtain the same retained-performance benefit.

\textbf{Directional evolution reduces the search component of the pipeline.}
Table~\ref{tab:pipeline-efficiency} reports 29--54\% fewer search rollouts for Harness-S, with GPU hours and API tokens within the range of the joint-adaptation baselines.
Relative to WHALE, Harness-S uses fewer GPU hours but more API tokens; relative to HarnessForge, it uses fewer API tokens but more GPU hours.
HarnessForge's zero additional training rollouts reflects reuse of successful search trajectories for policy updates.
Re-anchoring supports reuse across policy checkpoints; extending the pool between training runs can amortize earlier evolution effort.

\FloatBarrier
\section{Extended Related Work}
\label{app:extended-related-work}

\subsection{Agent RL Training}
\label{app:agent-rl-training}

Interactive environments make action selection and observation handling part of the training problem.
Toolformer learns API use through self-supervised examples, while AgentTuning initializes agent capabilities through instruction tuning on interaction trajectories~\cite{schick2023toolformerlanguagemodelsteach,zeng2023agenttuningenablinggeneralizedagent}.
Such supervision can initialize tool-use behavior before reinforcement learning from task outcomes.
WebGPT combines browsing demonstrations with human-preference feedback~\cite{nakano2022webgptbrowserassistedquestionansweringhuman}.
ReTool learns to interleave reasoning with code execution, while Search-R1 learns multi-turn search interactions and masks retrieved content during optimization~\cite{retool,jin2025searchr1trainingllmsreason}.
SWE-RL develops software-engineering reasoning from software evolution data with rule-based solution rewards~\cite{swerl}.
The supervision source and action interface jointly determine which behaviors can be learned.

Long interactions introduce credit-assignment and rollout-quality problems beyond optimizing individual responses.
RAGEN identifies instability associated with collapsing reward variation and studies how rollout diversity and feedback shape multi-turn learning~\cite{ragen}.
LOOP trains agents directly in stateful environments such as AppWorld using a memory-efficient PPO variant~\cite{chen2025reinforcementlearninglonghorizoninteractive}.
AgentGym-RL explores a complementary curriculum, gradually increasing interaction horizons to balance early exploitation with broader exploration~\cite{xi2025agentgymrltrainingllmagents}.
Agent Lightning separates execution from optimization and decomposes trajectories into transitions with assigned credit~\cite{lightning}.
These directions distinguish the learning objective, interaction curriculum, and runtime interface through which training experience is collected.

Classical RL offers additional ways to learn when successful outcomes are rare or delayed.
Hindsight Experience Replay reuses unsuccessful attempts by relabeling them with goals that were achieved~\cite{andrychowicz2018hindsightexperiencereplay}.
RUDDER redistributes delayed rewards through return decomposition~\cite{arjonamedina2019rudderreturndecompositiondelayed}, whereas Go-Explore separates returning to promising states from exploring beyond them~\cite{ecoffet2021goexplorenewapproachhardexploration}.
These methods intervene in goal assignment, credit distribution, and state coverage, respectively.
The distinction matters for language agents: changing how experience is generated and changing what an existing trajectory teaches are different interventions.

Uniformly rewarded groups limit the signal available to group-relative updates.
PPO uses a surrogate objective to control policy updates, while GRPO estimates advantages from rewards within a sampled group~\cite{ppo,deepseekmath}.
DAPO uses dynamic sampling to retain groups with reward variation, filtering groups whose samples are all correct or all incorrect~\cite{yu2025dapoopensourcellmreinforcement}.
Model-aware tool-use data evolution in LoopTool and hint-conditioned generative recommendation in HCGRec address related data and exploration bottlenecks~\cite{zhang2026looptool,zhang2026hcgrec}.
Harness-S changes execution support before generating fresh attempts on the same failed task.
It keeps the task and verifier fixed while seeking the successful alternatives needed for policy updates.

\subsection{Knowledge Transfer}
\label{app:knowledge-transfer-related}

The form of supervision determines what a student can inherit from a teacher.
Classical knowledge distillation transfers predictive distributions~\cite{hinton2015distillingknowledgeneuralnetwork}, while Sequence-Level Knowledge Distillation trains autoregressive students with teacher-generated sequences~\cite{kim2016sequencelevelknowledgedistillation}.
These approaches transfer information at different granularities: relations among candidate outputs versus complete generated solutions.
The distinction is relevant when successful behavior consists of a sequence of decisions rather than a single prediction.

Multiple specialists can also be consolidated into a shared decision-making model.
Policy Distillation combines task-specific reinforcement-learning policies into a single network~\cite{rusu2016policydistillation}.
Kickstarting combines guidance from previously trained agents with continued reinforcement learning, allowing a student to benefit from several experts and improve beyond them~\cite{schmitt2018kickstartingdeepreinforcementlearning}.
The resulting learner can act without querying every original expert, separating the resources used to acquire a capability from those required to deploy it.

Additional computation and reasoning traces provide another source of transferable knowledge.
Expert Iteration alternates tree-search planning with training a policy to generalize the resulting decisions~\cite{anthony2017thinkingfastslowdeep}.
STaR trains on successful self-generated rationales, including rationales regenerated with access to the correct answer after failure~\cite{zelikman2022starbootstrappingreasoningreasoning}.
Distilling Step-by-Step uses teacher-generated rationales as additional supervision alongside task labels~\cite{hsieh2023distillingstepbystepoutperforminglarger}.
These approaches differ in how they obtain useful traces, but all use richer training-time computation to shape a model's subsequent behavior.

Transfer must also account for the distribution induced by the learner.
DAgger aggregates expert-labeled states visited by successive learner policies to address distribution shift in sequential imitation~\cite{ross2011reductionimitationlearningstructured}.
Generalized Knowledge Distillation similarly obtains teacher feedback on student-generated sequences~\cite{agarwal2024onpolicydistillationlanguagemodels}.
MiniLLM studies on-policy optimization of reverse KL divergence to limit student probability mass on outputs that the teacher assigns low likelihood~\cite{gu2026minillmonpolicydistillationlarge}.
Together, these works show why the training distribution and the matching objective are separate design choices.

Teachers may additionally have information that the deployed learner will not receive.
Generalized distillation connects this setting to learning with privileged information~\cite{lopezpaz2016unifyingdistillationprivilegedinformation}; Learning by Cheating applies it to driving by training a vision-based policy from an agent with privileged environmental state~\cite{chen2019learningcheating}.
VAD examines the source of multimodal teacher feedback through controlled visual-evidence interventions~\cite{zhang2026vad}.
Harness-S instead obtains behavioral diversity by executing the same current policy under different harness contexts.
Task outcomes supervise its trajectories, and evaluation under $H_0$ tests whether useful behaviors persist when auxiliary execution support is removed.

\subsection{Harness Editing and Reusable Skills}
\label{app:harness-and-skills}

The execution interface determines how a language model's outputs become actions.
ReAct interleaves reasoning and environment interaction~\cite{react}; SWE-agent and CodeAct study dedicated computer interfaces and executable code actions, respectively~\cite{sweagent,codeact}.
AutoGen extends the design space to configurable conversations among agents, tools, and human participants~\cite{wu2023autogenenablingnextgenllm}.
These systems establish different control structures and action representations that later optimization procedures can modify.

Execution support can be optimized at several levels of granularity.
OPRO uses evaluated solutions to guide new prompt proposals~\cite{opro}.
EvoPrompt applies evolutionary operators to prompt populations, while Promptbreeder also evolves the mutation prompts that generate candidates~\cite{evoprompt,promptbreeder}.
DSPy compiles modular language-model pipelines against task metrics~\cite{dspy}.
TextGrad propagates textual feedback through computation graphs, and AFlow searches code-represented workflows with execution feedback~\cite{textgrad,aflow}.
AgentOptimizer treats an agent's functions as adjustable parameters while keeping the underlying language model fixed~\cite{zhang2024offlinetraininglanguagemodel}.
Thus, prompt editing, workflow search, and function revision expose different parts of an execution system to optimization.

Maintaining an archive supports exploration of multiple improvement directions.
ADAS searches code-defined agents using previous discoveries, and the Darwin Gödel Machine branches from an archive of executable agents whose modifications are evaluated empirically~\cite{adas,dgm}.
GEPA combines trajectory-based reflection with complementary prompt candidates on a Pareto frontier~\cite{gepa}.
Meta-Harness gives its proposer access to earlier implementations, scores, and execution traces~\cite{metaharness}.
Harness-R1 makes the editing policy itself learnable from the outcomes of its runtime changes~\cite{shao2026harnessr1}.
Archive design and proposal learning therefore address distinct parts of search: retaining alternative behaviors and improving how candidates are generated.

Reusable tools and skills reduce the need to rediscover useful procedures.
LLMs As Tool Makers separates a model that constructs reusable tools from one that applies them, allowing construction costs to be shared across later requests~\cite{cai2024largelanguagemodelstool}.
Voyager accumulates executable skills through interaction, while Agent Workflow Memory induces reusable routines from past experience~\cite{voyager,awm}.
The reusable object can consequently range from a callable function to a workflow supplied as context, with different requirements for retrieval and execution.

Natural-language memory offers another way to preserve experience without immediately updating model weights.
Reflexion stores linguistic feedback for subsequent trials, ExpeL extracts insights from past tasks, and ACE incrementally curates strategies in context playbooks~\cite{reflexion,zhao2024expelllmagentsexperiential,ace}.
MMSkills extends reusable packages to multimodal evidence; SkillRise and SkillGate learn skill curation and selection, respectively~\cite{zhang2026mmskills,yao2026skillriseagenticreinforcementlearning,li2026skillgatetraininginpolicyskill}.
Representing a procedure, deciding when to retrieve it, and retaining its benefits in policy parameters remain separate questions.

Execution resources can evolve jointly with the policy or be allocated during its training.
Co-Harness and WHALE alternate harness and model optimization, while HarnessForge evolves coupled harness--policy systems~\cite{coharness,whale,harnessforge}.
HarnessBandit instead schedules one harness per optimizer step using learnability and cross-harness transferability signals~\cite{harnessbandit}.
Harness-S allocates a fixed sibling pool at the task level, invoking auxiliary harnesses after all initial attempts under $H_0$ fail.
Policy-aware profiles track which variants remain useful, while rescue trajectories update one shared policy.
This links harness search to policy learning through complementary exploration, with base-harness evaluation separating retained policy gains from additional inference-time support.

\clearpage
\end{document}